\documentclass{acmsiggraph}
\usepackage{titlesec}

\pdfoutput=1

\titlespacing{\section}{0pt}{*0}{*0}
\titlespacing{\subsection}{0pt}{*0}{*0}
\titlespacing{\subsubsection}{0pt}{*0}{*0}

\title{Bringing Impressionism to Life with Neural Style Transfer in \textit{Come Swim}}

\author{Bhautik J Joshi\thanks{e-mail:bjoshi@gmail.com}\\Research Engineer, Adobe
       \and
       Kristen Stewart\\Director, \textit{Come Swim}
       \and
       David Shapiro\\Producer, Starlight Studios}
\pdfauthor{Bhautik J Joshi}


\CopyrightYear{2017}
\setcopyright{rightsretained}

\begin{document}


 \teaser{
   \includegraphics[height=1.5in]{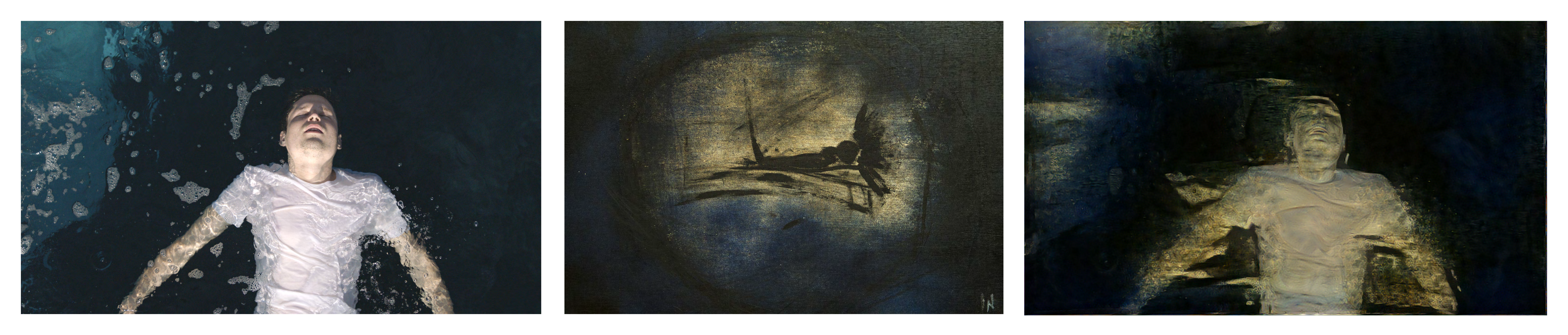}
   \caption{Usage of Neural Style Transfer in Come Swim; left: content image, middle: style image, right: upsampled result. Images used with permission, (c) 2017 Starlight Studios LLC \& Kristen Stewart.}
 }

\maketitle

\begin{abstract}

Neural Style Transfer is a striking, recently-developed technique that uses neural networks to artistically redraw an image in the style of a source style image. This paper explores the use of this technique in a production setting, applying Neural Style Transfer to redraw key scenes in \textit{Come Swim} in the style of the impressionistic painting that inspired the film. We document how the technique can be driven within the framework of an iterative creative process to achieve a desired look, and propose a mapping of the broad parameter space to a key set of creative controls. We hope that this mapping can provide insights into priorities for future research.

\end{abstract}

%
%


%
%


\keywordlist

\conceptlist

\printcopyright

\section{Introduction}
In \textit{Image Style Transfer Using Convolutional Neural Networks}, Gatys et al~\cite{gatys:2015} outline a novel technique using convolutional neural networks to re-draw a content image in the broad artistic style of a single style image. A wide range of implementations have been made freely available~\cite{liu:2016,johnson:2015,athalye:2015}, based varyingly on different neural network evaluators such as Caffe~\cite{jia:2014} and Tensorflow~\cite{abadi:2016} and wrappers such as Torch and PyCaffe~\cite{bahrampour:2015}. There has been a strong focus on automatic techniques, even with extensions to coherently process video~\cite{ruder:2016}. Workflows exist that permit painting and masking to guide the transfer more explicitly, as well as to reduce the size of the neural net for evaluation to significantly speed up execution~\cite{ulyanov:2016}. Creative control of these implementations are broadly handled via parameter adjustment.

Using the technique with a small set of carefully picked style examples is a natural optimization, allowing Neural Style Transfer to execute efficiently and predictably. In a production setting, however, a great deal of creative control is needed to tune the result, and a rigid set of algorithmic constraints run counter to the need for this creative exploration. While early investigations to better map the low-level neural net evaluations to stylistic effects are underway~\cite{li:2017}, in our paper we focused on examining the higher-level parameter space for Neural Style Transfer and found a set of working shortcuts to map them to a reduced but meaningful set of creative controls.

\section{Realizing Directorial Intent}
\begin{figure}[ht]
  \centering
  \includegraphics[width=3.0in]{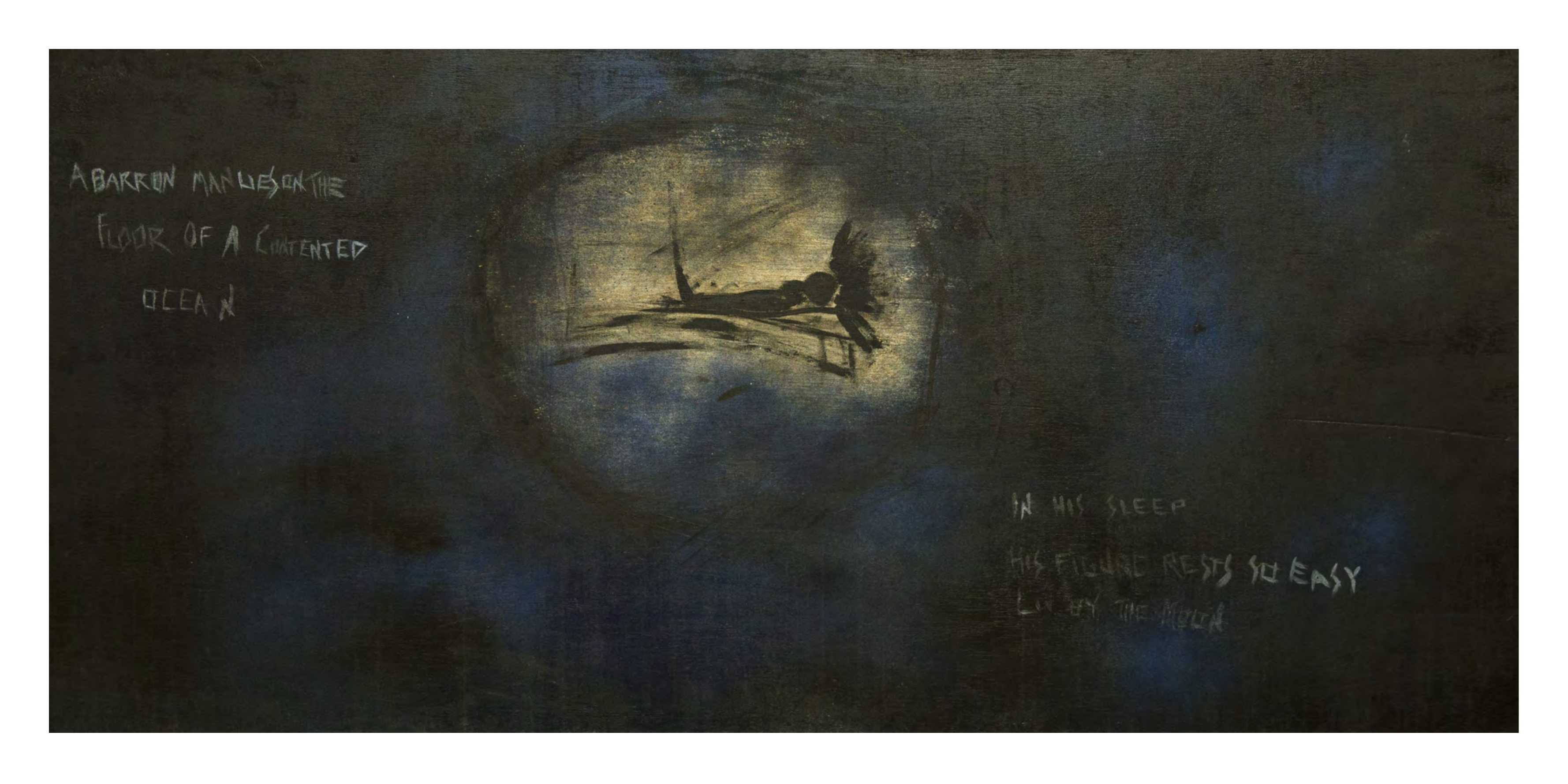}
  \caption{The painting that drove the story and aesthetic for \textit{Come Swim}. Image used with permission, (c) 2017 Starlight Studios LLC \& Kristen Stewart.}
  \label{fig:painting}
\end{figure}

\textit{Come Swim} is a poetic, impressionistic portrait of a heartbroken man underwater. The film itself is grounded in a painting (figure~\ref{fig:painting} by coauthor Kristen Stewart) of man rousing from sleep. We take a novel artistic step by applying Neural Style Transfer to redraw key scenes in the movie in the style of the painting, realizing them almost literally painting that underpins the film.

The painting itself evokes the thoughts an individual has in the first moments of waking (fading in-between dreams and reality), and this theme is explored in the introductory and final scenes where this technique is applied. This directly drove the look of the shot, leading us to map the emotions we wanted to evoke to parameters in the algorithm as well as making use of more conventional techniques in the 2D compositing stage.

\section{Initial Setup}
Initial development of the look on the shots was arguably the longest and most difficult part of the process. The novelty of the technique gave a false sense of a high-quality result early on; seeing images redrawn as paintings is compelling enough that nearly any result seems passable. Repeated viewings were needed to factor out novelty and to focus on the primary metric we use to measure success - is the look in service of the story?

\subsection{Pipeline for Evaluation}
Given that the technique has not been used often in formal VFX production, we had to establish a baseline. The parameter space and corresponding creative possibilities are huge, so initial iteration was focused on shrinking down the number of possibilities down to relevant options.

Whilst the technique appears to deliver impressionism on demand, steering the technique in practice is difficult. We examined directable versions of the technique~\cite{ulyanov:2016} but time pressures meant that we were not tooled for this approach and preferred the whole-frame consistency - and serendipity - that came from applying the technique uniformly to the entire frame. We focused on techniques we could prototype locally and with modest hardware~\cite{liu:2016}. Strong coherency between frames~\cite{ruder:2016} was not desired creatively for the sequences considered, which permitted an extra degree of latitude for developing the look.

For the neural network evaluation, we initially tried the \texttt{googlenet}~\cite{szegedy:2015} and \texttt{vgg19}~\cite{simonyan:2014} neural nets; \texttt{vgg19} execution times were too high, and \texttt{googlenet} did not give the aesthetics we were seeking. We made an early choice to settle on \texttt{vgg16}~\cite{simonyan:2014} since it provided the right balance between execution time and the quality of result we desired.

Early results were achieved by working on a single, representative frame to iterate on the look. Its also worth noting the near linear relationship between style transfer ratio and image size; working with an image that has half the image dimensions meant roughly halving the style transfer ratio. Faster, coarse iteration could then be performed on lower-resolution frames.

\subsection{Iterating on the style image}
\begin{figure}[ht]
  \centering
  \includegraphics[width=3.0in]{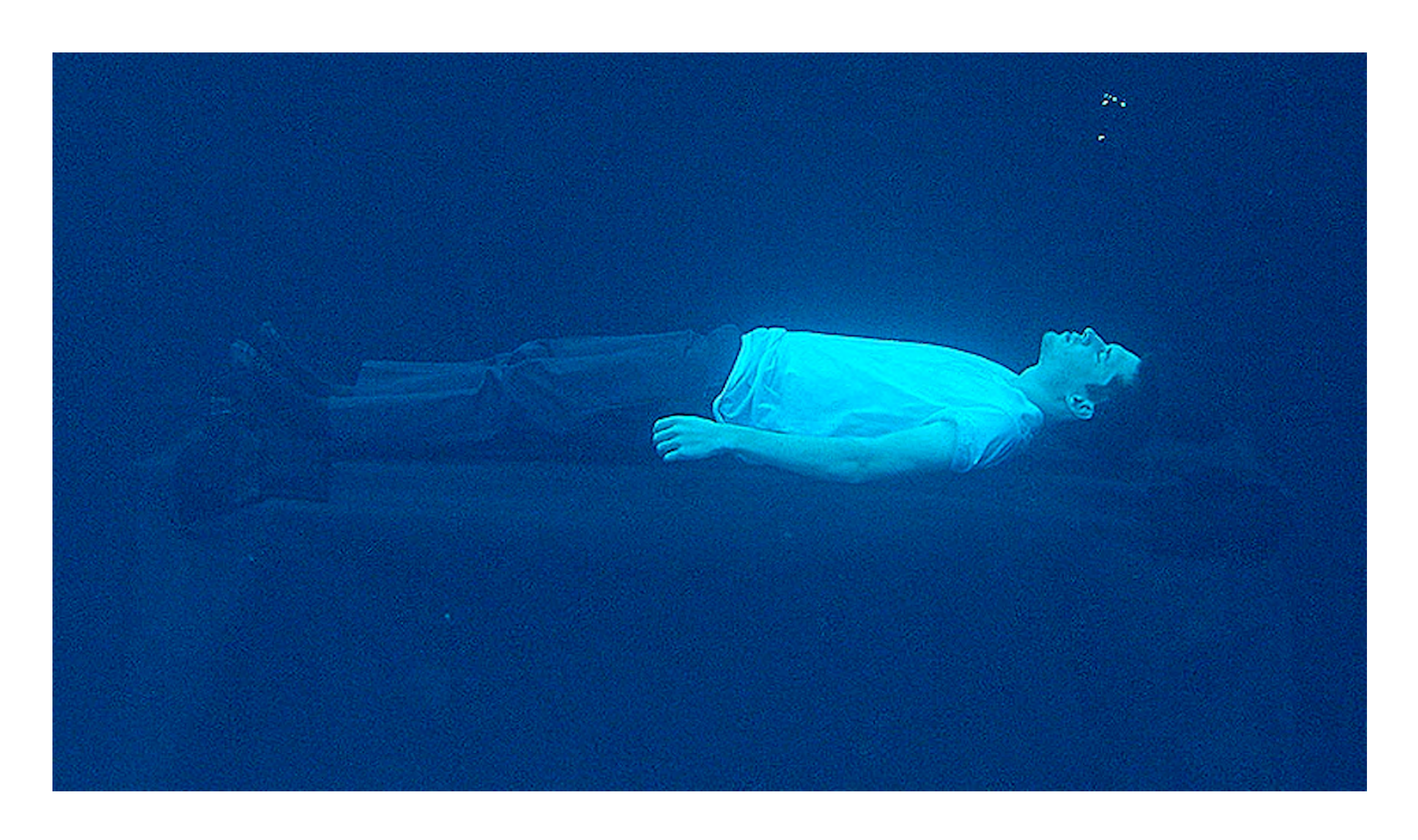}
  \caption{The reference image used for during tuning of the style image. Image used with permission, (c) 2017 Starlight Studios LLC \& Kristen Stewart.}
  \label{fig:reference_content}
\end{figure}

An early observation during work on the setup of the style transfer was that the effect could be directed by changing the content of the style image itself. For the iterations presented here, we used a fixed content image as shown in figure~\ref{fig:reference_content}.

\begin{figure}[ht]
  \centering
  \includegraphics[width=3.0in]{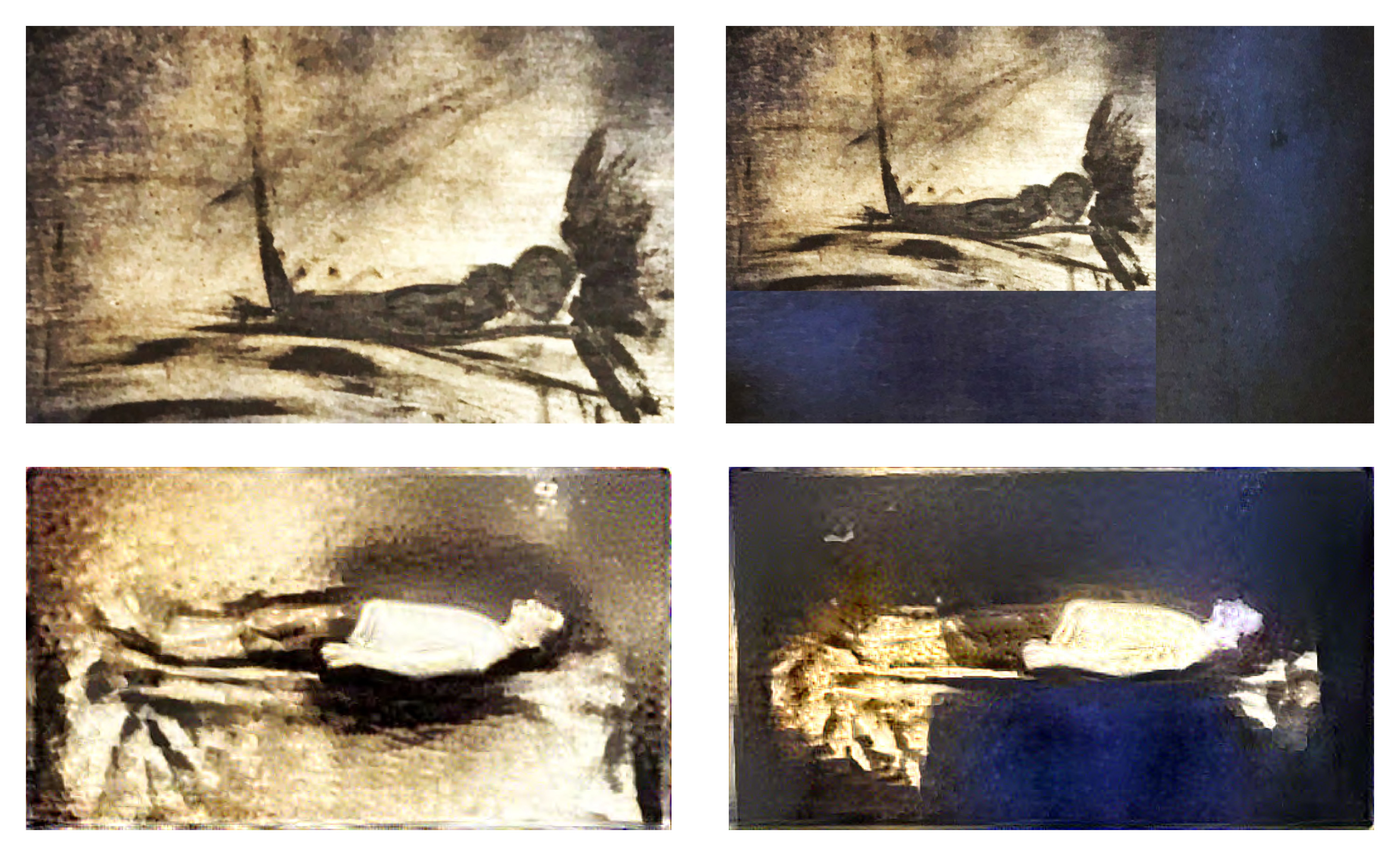}
  \caption{Initial style image (left-top) and style transferred result (left-bottom). Larger blocks of the style image were added (right-top) generating a richer result (right-bottom).}
  \label{fig:styletransfer_A}
\end{figure}

We initially controlled for color and texture by closely cropping the style image. This set up a baseline for the look of texture transfer (figure~\ref{fig:styletransfer_A}, left). We experimented with the number of iterations in the algorithm, optimizing for speed to see how much it caused the look to deviate, giving us a reasonable range for this parameter. We added larger blocks of color and texture to the style image (figure~\ref{fig:styletransfer_A}, right). As the texture is built, other parameters for the style transfer can be more finely tuned and brought into a smaller, more manageable range for iteration on look.

\begin{figure}[ht]
  \centering
  \includegraphics[width=3.0in]{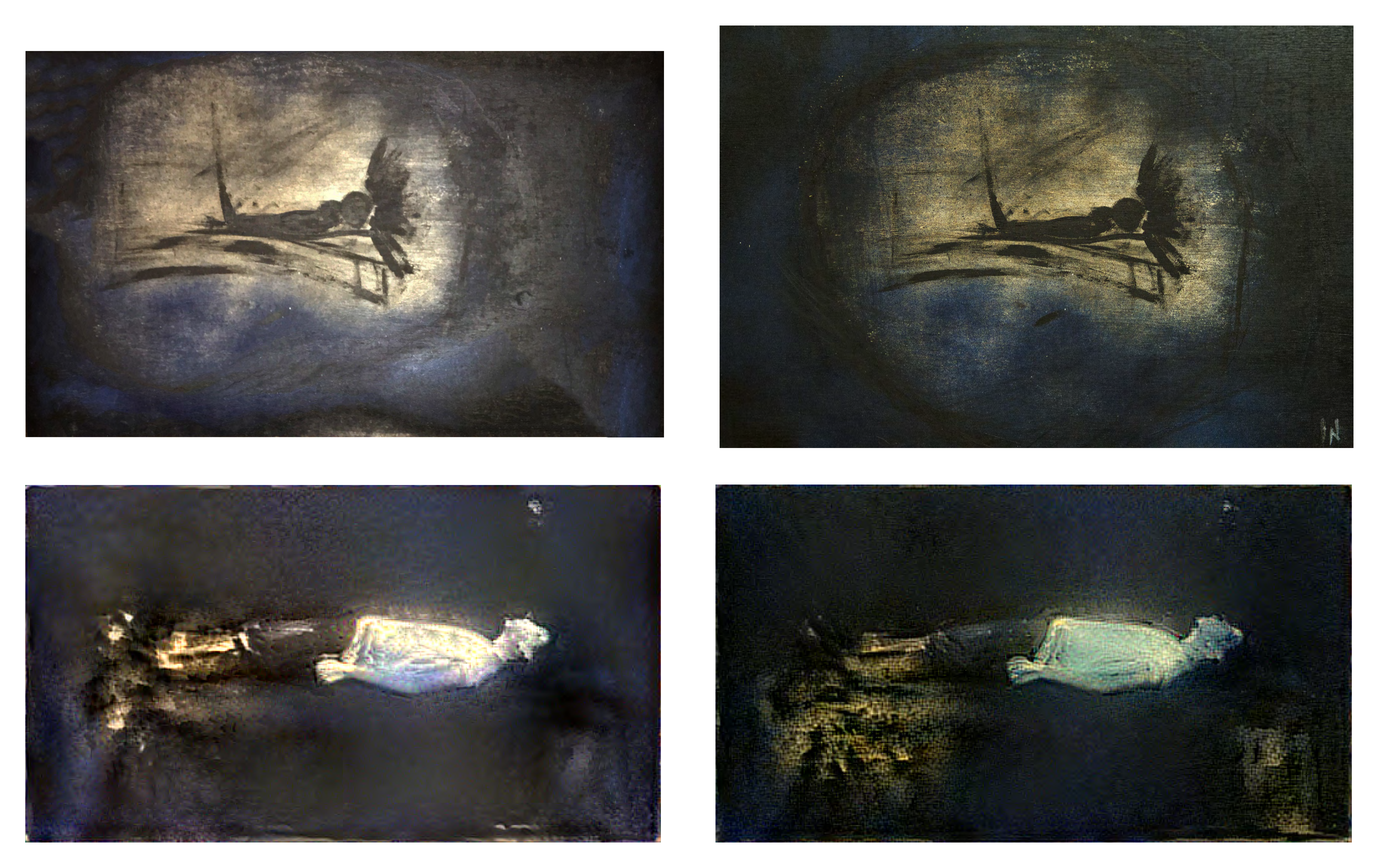}
  \caption{The quality of the style image is critical - a low-quality photograph of the painting (top-left) leads to poor transfer of texture and contrast (bottom-left); a higher quality photograph (top-right) leads to better detail and texture preservation (bottom-right).}
  \label{fig:styletransfer_B}
\end{figure}

The quality of the input style image is also critical for achieving a high quality result. The painting that we used for the style image is texturally complex - the paint itself is reflective and is applied in high-frequency spatter across the canvas. A low-quality cellphone shot of the canvas drove initial work on the shots (figure~\ref{fig:styletransfer_B}, left). This low-quality image blew out the reflective highlights and flattened out the subtle texture on the paint, and this same artifact is can be seen in the style transfer. Replacing the style image with a well-lit, high-quality photograph significantly improves the result, allowing for more faithful transfer of the subtleties in the contrast, color and texture (figure~\ref{fig:styletransfer_B}, right).

\subsection{Mapping look to parameters}
Once the pipeline was initialized, we identified two major parameters as it applied to our look - the style transfer ratio and the number of iterations. Increasing the number of iterations increases the degree of texture that is transferred; in general it needs to be at least 128 to prevent noticeable artifacts in the content image. We found that in practice 256 iterations were sufficient for the style image we were using; much past this only produced negligible improvements in (the subjectively assessed) quality of the style transfer.

A fixed number of iterations, led to a near-constant execution time per-frame. Experimenting with the style transfer ratio led us to conclude that it needed to be exponent form for meaningful creative exploration. Subjectively, this exponential form gave us a useful measure of unrealness, \textit{u}, a rough way to map how impressionistic the style transferred image looked:

\begin{equation}
style\ transfer\ ratio = 10^u
\end{equation}

\begin{figure}[ht]
  \centering
  \includegraphics[width=3.0in]{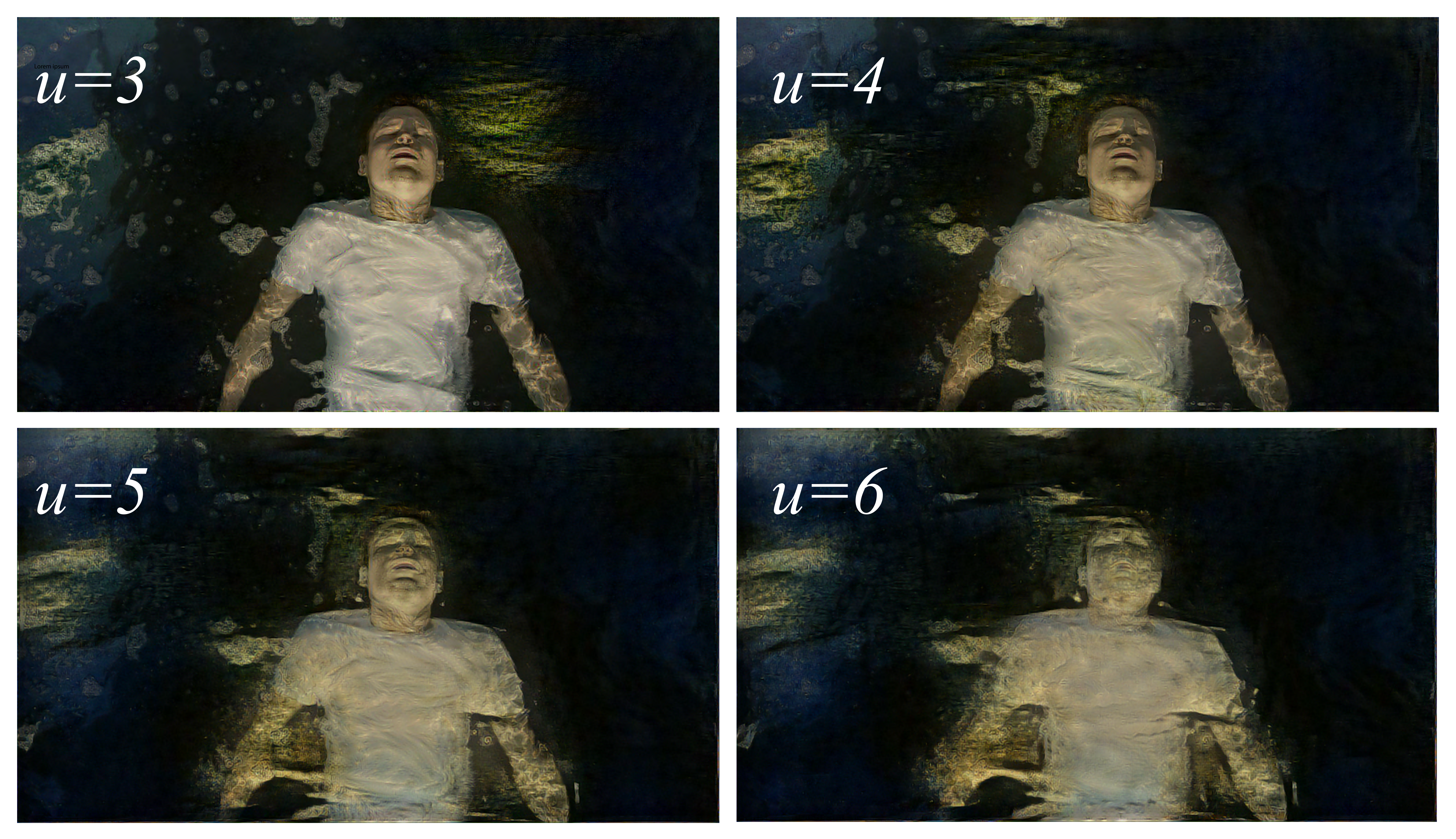}
  \caption{Increasing the value of \textbf{u} gave us a control to fine-tune the degree of subjectively-measured impressionism present in the style transfer}
  \label{fig:parameter_space}
\end{figure}

As shown in figure~\ref{fig:parameter_space}, this mapping allowed us to iterate on the intensity of the style transfer. To achieve the artistic effect desired and to magnify the effect artistically, the style-transferred footage was cross-dissolved in and out of the original footage.

\section{Production Notes}
As mentioned earlier, we used low-resolution previews computed locally on modest hardware on the CPU to set up the production pipeline. However, to get us to the needed level of quality we needed a scalable solution that provided GPU resources on demand - in this case, we chose to use GPU instances on Amazon EC2.

We used custom ubuntu images on \texttt{g2.2xlarge} virtual instances to provide us with the necessary computing power to achieve the result in a reasonable amount of time. However, the process itself is fairly heavily limited by the amount of available GPU memory; in the configuration we used, we were realistically only able to produce 1024px wide images; the show called for 2048px images.

To work around this, we upscaled the images to 2048px resolution and denoised the images aggressively. This functioned as a type of bandpass filter, and worked well to achieve the sketch-based look we were after. Overall, we ended up with a compute time of around 40 minutes per frame per instance used.

\section{Conclusion and Future Explorations}
Far from being automatic, Neural Style Transfer requires many creative iterations when trying to work towards a specific look for a shot. The parameter space is very large - careful, structured guidance through this in collaboration with the creative leads on the project is needed to steer it towards a repeatable, visually pleasing result.

Gradually building the style image in terms of texture and color blocks to move towards the desired result can help frame the process and set baselines for look and execution time. Quick iteration can also be achieved by using low-resolution images with appropriately scaled style transfer ratios allows for representative results before fine-tuning with much longer-running, high-resolution iterations.

An initial future optimization that would seem obvious first would be to use an optimized, feed forward model~\cite{ulyanov:2016}. However, the single style transfer ratio proposed in such models would not meet the needs for flexibility around this parameter for creative iteration in shots. A more promising approach would be to a pastiche-based style transfer model~\cite{dumoulin:2016}, but instead of using several different style images, use the same style image with a spread of style transfer ratios.

\bibliographystyle{acmsiggraph}
\nocite{*}
\bibliography{digipro17}
\end{document}